\documentclass{article}

\usepackage{stfloats}
\usepackage[utf8]{inputenc}
\usepackage{adjustbox}
\usepackage{geometry}
\usepackage{pdflscape}
\usepackage{amsthm}
\usepackage{amsmath}
\usepackage{amssymb}
\usepackage{listings}
\usepackage{parskip}
\usepackage{graphicx}

\usepackage{biblatex}
\begin{document}
%
\title{Perfect Edge-Transmitting Recombination of Permutations}
%
%
%
\author{Adriaan Merlevede and Carl Troein}
\author{Adriaan~Merlevede~
        and~Carl~Troein%
\thanks{A. Merlevede and C. Troein are with the Department of Astronomy and Theoretical Physics, Lund University, Lund, Sweden}%
}

%
%

\markboth{Preprint}{To be submitted}
%



\maketitle

\paragraph{Abstract}
Crossover is the process of recombining the genetic features of two parents. For many applications where crossover is applied to permutations, relevant genetic features are pairs of adjacent elements, also called edges in the permutation order. Recombination of edges without errors is thought to be an NP-hard problem, typically approximated by heuristics that either introduce new edges or are only able to produce a small variety of offspring. Here, we derive an algorithm for crossover of permutations that achieves perfect transmission of edges and produces a uniform sampling of all possible offspring, in quadratic average computation time. The algorithm and its derivation reveal a link between cycle crossover (CX) and edge assembly crossover (EAX), offering a new perspective on these well-established algorithms. We also describe a modification of the algorithm that generates the mathematically optimal offspring for the asymmetric travelling salesman problem.



\section{Introduction}

The recombination of genetic material occurring in sexual reproduction is an important process that greatly affects evolutionary dynamics. Recombination allows individual mutations to undergo natural selection separately, so that beneficial genotypic characteristics can be selected independently from other (potentially deleterious) genetic features. In addition, different beneficial mutations do not form separate and competing sub-populations, but can be combined. The process of recombination, also called crossover, can be applied to abstract data structures, allowing sexual reproduction in simulated evolutionary processes such as evolutionary algorithms. Crossover of permutations (orderings of a specific set of elements) is a particularly well-studied area.



Crossover strategies for permutations can be classified based on the characteristics of the parents that they attempt to recombine. Depending on context, crossover strategies are designed to consider the absolute position of symbols, the relative order of pairs, or the adjacency of elements. For example, a crossover based on adjacency attempts to produce offspring where pairs of adjacent symbols in the offspring are also adjacent in one of the parents. 
The objective of recombination can be formulated more rigorously using a similarity measure to define correct offspring as ``in between'' parents in a geometric sense\cite{moraglio_topological_2004,moraglio_geometric_2011}; or by utilising the concept of a schema or forma to mathematically define the relevant characteristic\cite{chen_commonality_1996,radcliffe_algebra_1994}.


Adjacency-based crossover has seen extensive application for graph-based optimisation problems. For example, the classic representative of this problem class is the travelling salesperson problem (TSP), where the task is to plot a tour through a number of cities while minimising the total travelled distance; in graph terms, to find the minimal-cost Hamiltonian cycle through a complete weighted graph. For such problems, pairs of adjacent elements in the cyclic ordering of visited nodes represent edges in the graph, which are the building blocks that determine the quality of a tour. A crossover that recombines edges produces offspring consisting of contiguous sections inherited from the parent tours. Depending on whether the cost associated with an edge or solution is dependent on its direction, we distinguish between directed and undirected edges, or equivalently between symmetric or asymmetric graph problems. While recombination of undirected edges may include a mixture parental edges in both directions (thus preserving adjacency of elements in the permutation), asymmetric graph problems demand that the orientation of edges is consistent (preserving adjacent pairs in the same order), which limits the variety of allowed offspring.

The usefulness of edge-based crossover strategies has led to a wealth of algorithms being published in the past several decades. Some are highly efficient at producing high-quality offspring for heuristic optimisation of certain problems such as the TSP, but when considering theoretical descriptions of edge-based recombination, most crossover strategies adhere only imperfectly to the established goals. Radcliffe et al. define three objectives of recombination, which we can apply to edge-based crossover: (\textit{respectful}) any offspring inherits all edges which are common to both parents, (\textit{transmissive}) all edges in any offspring occur in at least one of the parents, and (\textit{assorting}) any possible combination of the parents' edges can be inherited by some offspring\cite{radcliffe_fitness_1995}. With this in mind, the edge exchange crossover (EXX), the (generalized) partition crossover ((G)PX), and the asymmetric variant GAPX\cite{whitley_hybrid_2010,veerapen_tunnelling_2016} are notable because they adhere to the transmissive property. However, they are unable to produce all possible offspring that have only edges from the parents.

We will refer to a crossover that produces a uniform sampling of all possible offspring under the constraints of \textit{respectful} and \textit{transmissive} crossover as \textit{perfect}. In the case of edge-based crossover, \textit{perfect} crossover is as close as possible to \textit{assorting} without breaking the transmissive property\cite{radcliffe_fitness_1995,radcliffe_algebra_1994}. This concept of \textit{perfect} edge-based crossover coincides with what has previously been described as uniform geometric crossover for adjacency distance (AX)\cite{moraglio_topological_2004,moraglio_geometric_2011}. 


An implementation of \textit{perfect} edge-based crossover has not yet been described in literature. On the contrary, the problem has been described as unlikely to be solved by an efficient algorithm, because closely related problems in sorting and distance computation are NP-hard\cite{moraglio_geometric_2011,nagata_criteria_2004}. This is in contrast to other kinds of \textit{perfect} crossover for which efficient implementations are known, such as recombination based on absolute position (cycle crossover\cite{oliver_study_1987}) and relative order (merge crossover\cite{moraglio_geometric_2011,blanton_multiple_1993}). For heuristic optimisation, some authors have also argued that errors in crossover are preferable over perfect transmission, on the grounds that less disruptive crossover algorithms also have a lower variety of outcomes, which has the potential to result in premature convergence\cite{nagata_criteria_2004}.  On the other hand, \textit{transmissive} crossover has the useful property that when two parents in different local optima are recombined, offspring are likely located in a third local (or possibly global) optimum. The authors of G(A)PX call this ``tunneling'' between remote parts of the fitness landscape, which is particularly useful in conjunction with local search heuristics\cite{whitley_hybrid_2010,veerapen_tunnelling_2016}. There are also other situations where introduction of new edges is especially undesirable, for example in a variant of the TSP where the graph is sparsely connected.

In addition, guaranteed error-free crossover opens up new potential applications outside of heuristic optimisation, where recombination errors may not be tolerable. For example, in biology, changes to genome structure are a major component of natural variation within and between species, and play a critical role in speciation and other aspects of evolution. These structural changes include variation in the ordering of genetic components which occur through translocations and inversions, where contiguous segments of a chromosome are moved or reversed. Error-free edge-based recombination can be used to investigate potential ways to recombine such structural changes while otherwise maintaining the same genomic architecture.


We derive from first principles an algorithm for \textit{perfect} crossover based on directed edges. The proposed algorithm is related to two previously published crossover strategies; the cycle crossover (CX), and the asymmetric version of the edge assembly crossover (EAX), and reveals a link between the two that simplifies the interpretation and implementation of the asymmetric EAX. We also propose a symmetric variant of the algorithm inspired by the symmetric EAX, and an alternative implementation that guarantees generation of the optimal offspring with respect to the asymmetric TSP. For each proposed algorithm, we analyse the computational requirements.

\section*{Background}

Formally, a permutation is a bijective function that maps the set $\{1,\dots,n\}$ to itself, so that a permutation $\sigma$ represents the ordering $\sigma(1), \sigma(2), \dots$. In the rest of this paper, we use product notation for the left-to-right composition of permutations, defining $(a \cdot b)(i) \equiv (ab)(i) \equiv b(a(i))$, which is common in permutation theory. We also assume a cyclic interpretation of permutations, so that we define $\sigma(n+1) \equiv \sigma(1)$ and $\sigma(0) \equiv \sigma(n)$, and include the edge $\sigma(n) \rightarrow \sigma(1)$ where applicable. Our focus on cyclic permutations does not result in loss of generality because non-cyclic problems (such as the non-cyclic TSP) can be expressed in terms of cyclic permutations by adding an extra node to represent the start of the sequence.

\subsection*{Cycle crossover (CX)}
Cycle crossover is one of the first proposed crossover strategies for permutations\cite{oliver_study_1987}. It is \textit{transmissive} with respect to absolute position of elements, and its implementation follows directly from a simple and elegant argument. It is also \textit{respectful} and able to produce all possible offspring (within the transmissive constraint) with equal probability, and is therefore a \textit{perfect} crossover.

We present briefly the derivation of cycle crossover. An offspring $c$ that is a recombination of parents $a$ and $b$ may have in position $i$ only one of the symbols $a(i)$ or $b(i)$. Let us assume that $c$ inherits $a(i)$ in position $i$. Then, $c$ cannot inherit $b(i)$ in position $i$, and therefore must inherit $b(i)$ from parent $a$, thus in position $j=a^{-1}(b(i))$. By the same reasoning we must also have that $c$ cannot inherit $b(j)$ from parent $b$, and thus inherits $b(j)$ also in position $a^{-1}(b(j)$. Continuing this chain of reasoning, we conclude that the symbols in positions $i,\pi(i),\pi(\pi(i)),\dots$ have linked inheritance, where $\pi=b \cdot a^{-1}$. These linked families $\{ \{i,\pi(i),\pi(\pi(i)),\dots\} |\; i \in \{1,\dots,n\} \}$ are called the cycles of $\pi$ in permutation theory.

Concretely, we have that for any cycle $k$ of $\pi$, the offspring $c$ inherits either $c(i)=a(i)$ for all $i\in k$, or alternatively $c(i)=b(i)$ for all $i\in k$. This is enough to implement cycle crossover: compute the cycles of $\pi=b\cdot a^{-1}$, then choose for each cycle $k$ whether it will be inherited by $a$ or $b$ and construct the offspring $c$ accordingly. Note that cycles $\{i\}$ consisting of only one element represent elements that are in the same position in the two parents. These singleton cycles can be ignored because whether they are inherited from parent $a$ or $b$ has no influence on the outcome. 

Any position-based \textit{transmissive} crossover is also \textit{respectful}, because an element that is in the same position in the two parents leaves no alternative to be inherited the offspring. Cycle crossover produces all possible combinations of non-singleton cycles with the same probability and therefore results in a uniform sampling of all possible offspring subject to position-based transmission.

\subsection*{Edge assembly crossover (EAX)}

The edge assembly crossover is one of the most successful crossover strategies, in terms of efficiency for optimising solutions to the TSP in evolutionary algorithms. It is designed to combine transmission of most parent edges with a high variety of possible offspring\cite{nagata_edge_1997}. Newly introduced edges are optimised for fitness, preferring edges with lower cost\cite{nagata_edge_1997}. We briefly restate the algorithm here, but it is explained in more detail elsewhere by its original author\cite{nagata_edge_1997,nagata_criteria_2004}.

The EAX is based on a concept called an AB-cycle, which is a set of edges originating from the two graphs represented by the parents of the crossover, $G_a$ and $G_b$. An AB-cycle forms a cyclic path through some of the nodes, wherein edges originating from $G_a$ are alternated with edges originating from $G_b$. The edges in the AB-cycle originating from $G_a$ or $G_b$ connect the same set of nodes. Therefore, the edges originating from $G_a$ in an AB-cycle are interchangeable with the edges originating from $G_b$, in the sense that they do not alter the number of edges connecting to any of the nodes. In particular, using the AB-cycles to exchange edges in $G_a$ or $G_b$, which are 2-regular graphs meaning that each node is connected to precisely two edges, results in another 2-regular graph. The algorithm uses this property to recombine edges from the two parent graphs:

\begin{enumerate}
\item Construct the graphs $G_a$ and $G_b$ from the parent permutations, and denote the union graph $G_{ab}$
\item Divide the edges of the union graph $G_{ab}$ into AB-cycles.
\item Construct an E-set, which is a union of AB-cycles, according to a given rule, such as choosing a uniformly random union.
\item Generate an intermediate solution by removing all edges in the E-set from $G_a$, and replacing them with the edges in the E-set that originate from $G_b$.
\item Modify the intermediate solution to generate a valid tour.
\end{enumerate}

The last step is necessary because using an E-set to recombine edges in the parent graphs is guaranteed to result in a new 2-regular graph, but not necessarily a valid tour of all nodes. A 2-regular graph can consist of multiple disconnected tours which each visit a subset of all nodes. Thus, in the last step, the connected components of the intermediate solution are identified and merged. This procedure requires the introduction of new edges, which are not found in either parent. The new edges are chosen carefully to maximise the fitness of the offspring, with respect to a TSP-like fitness function.

The EAX can be applied for either a directed or undirected interpretation of the edges by using slightly difference procedures to generate AB-cycles in step 2. In the directed (asymmetric) variant, the directions of the edges are retained, but the edges of one of the parent graphs are reversed. In the undirected variant, the direction of the edges is forgotten when constructing the undirected union graph. In both cases, the AB-cycles are found by a greedy algorithm, which starts at a random node and traverses the edges originating from $a$ and $b$ in alternating fashion. If there is more than one option, which is only possible in the symmetric variant, one is chosen randomly, but each edge is traversed only once. Whenever the path generated by this procedure generates a loop of even length (thus containing the same number of edges from $a$ and $b$), the edges of that loop are taken out of the stored path, and labelled as an AB-cycle. When the algorithm is stuck in a node that does not connect to unvisited edges, it is continued at another random node. The algorithm finishes when all nodes have been assigned to an AB-cycle.\cite{nagata_edge_1997}

\subsection*{Adjacency representation}

A tour of all nodes in a graph can be represented in a straightforward manner as a permutation by listing all nodes in the order they are visited. This is called the path-representation of a tour. However, other mappings between tours and permutations are possible and have been used in genetic algorithms\cite{michalewicz_genetic_1996,radcliffe_fitness_1995}. In the adjacency-representation, each permutation element represents an edge of the tour: an edge $i \mapsto j$ is represented in the permutation $E_\sigma$ by $E_\sigma(i)=j$. In other words, the permutation maps each node to the next node in the tour. The following two tours are equivalent, for appropriate choices of $\sigma$ and $E_\sigma$:

\makebox[2em]{}path-interpretation of $\alpha$:\\[.5ex]
\makebox[3em]{}
\makebox[2em][c]{$\alpha(1)$}
\makebox[2em][c]{$\rightarrow$}
\makebox[2em][c]{$\alpha(2)$}
\makebox[2em][c]{$\rightarrow$}
\makebox[4em][c]{$\alpha(3)$}
\makebox[2em][c]{$\rightarrow$}
\makebox[2em][c]{$\dots$}  

\makebox[2em]{}adjacency-interpretation of $E_\alpha$:\\[.5ex]
\makebox[3em]{}
\makebox[2em][c]{$i$}
\makebox[2em][c]{$\rightarrow$}
\makebox[2em][c]{$E_\alpha(i)$}
\makebox[2em][c]{$\rightarrow$}
\makebox[4em][c]{$E_\alpha(E_\alpha(i))$}
\makebox[2em][c]{$\rightarrow$}
\makebox[2em][c]{$\dots$} 

if $i=\alpha(1)$, and indeed for any $i$ because the starting point of a cyclic tour is arbitrary.

This equivalence illustrates how an equivalent path-representation $\sigma$ can be found for an adjacency-permutation $E_\sigma$, by recording the nodes found by repeatedly applying $E_\sigma$. Conversely, for each $i$ we have $E_\sigma(\sigma(i))=E_\sigma(\sigma(i+1))$, which can be written as $E_\sigma = \sigma^{-1}\cdot r \cdot \sigma$, defining the rotation $r:i\mapsto i+1$.

Comparing these two representations, we note that the path-representation has redundancy because it encodes an arbitrary starting point in a cyclic tour, so that there are $n$ (the permutation size) permutations representing the same tour. The adjacency-representation does not suffer from this redundancy, there is precisely one permutation for each possible tour. However, only a fraction $1/n$ of all possible permutations represent a valid tour. Indeed, permutations with more than one cycle will visit only a part of the graph when applied repeatedly to a given starting point.

The path- and adjacency-representation are formulated here in terms of graph tours, or solutions to the TSP, but their use throughout the rest of this paper is not limited to the TSP. These concepts can be thought of more generally as an isomorphism, or mathematical equivalence, between the non-cyclic permutations $E_\sigma$ with one cycle, and equivalence classes of cyclic permutations $\sigma$ (cyclic here meaning that the ordering has no starting point).

\section*{Perfect crossover for directed edges}

\subsection*{Derivation of the algorithm}

Consider the constraints of \textit{transmissive} crossover for directed edges. The offspring $c$ of two parents $a$ and $b$ must consist of edges from $a$ and $b$ and be a valid permutation. In the adjacency representation, we have for each element $i$, that $c$ can have either the edge $i \mapsto E_a(i)$ or $i \mapsto E_b(i)$; assuming $E_a(i) \ne E_b(i)$ it is not possible to inherit both, or neither (if they are equal than the offspring must always inherit that edge). In other words $E_c(i)=E_a(i)$ or $E_c(i)=E_b(i)$. 

This simple argument reformulates the problem of (directed) edge-preserving crossover as a position-preserving crossover on adjacency representations. As a consequence, edge-preserving crossover is subject to the constraint of cycle crossover, namely that the cycles of the characteristic permutation $\pi$ (here equal to $E_bE_a^{-1}$) have linked inheritance. However, not all cycle crossovers of $E_a$ and $E_b$ result in a valid adjacency-representation $E_c$. Recall that a permutation is only a valid adjacency-representation of a tour if it has only one cycle, which is not necessarily true for all position-tranmissive recombinations of $E_a$ and $E_b$.

\begin{figure}[tb!]
    \centering
    \includegraphics[width=.6\linewidth]{{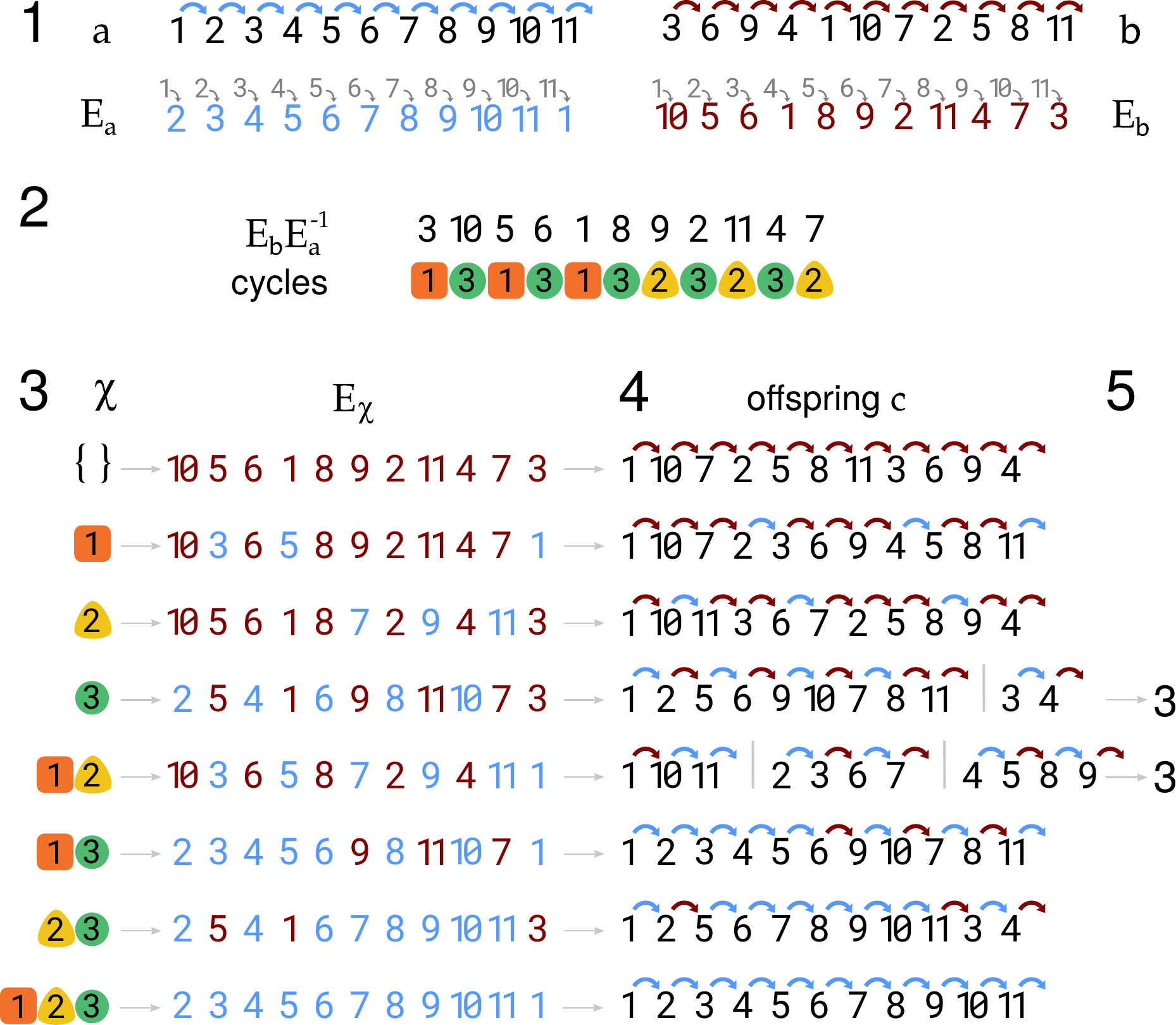}}
    \caption{Illustration of the asymmetric crossover algorithm. The figure is divided in five steps, corresponding to the five steps of the algorithm as detailed in the main text. Permutation edges are coloured blue or red by parent.}
    \label{fig:schematic}
\end{figure}

It is difficult to identify precisely the combinations of cycles that can be exchanged to produce valid adjacency-representations of offspring. Instead, since cycle crossover is computationally cheap, we propose a simple approach where the cycle crossover is repeated until a valid offspring is obtained. The resulting full algorithm consists of five steps illustrated in Fig.~\ref{fig:schematic}:

\begin{enumerate}
\item Calculate the adjacency representations $E_a=a^{-1}ra$ and $E_b=b^{-1}rb$ of the parents $a$ and $b$.
\item Identify the cycles of $E_b E_a^{-1}$.
\item Pick a union of cycles $\chi$.
\item Follow the cycle order of $E_\chi$ with $E_\chi(i)= E_a(i)\text{ if }i\in\chi\text{ else }E_b(i)$ starting from $1$ (starting point is arbitrary) to produce $c$.
\item If $c$ is a full-length offspring, finish. Otherwise, repeat from step $3$.
\end{enumerate}

The permutation $E_\chi$ is the cycle crossover of $E_a$ and $E_b$. In step 2 or 3, the cycles $\{i\}$ of size one may be ignored, as they do not influence $E_\chi$. This crossover algorithm is \textit{respectful} and uniformly samples all possible offspring subject to directed edge-preserving transmission, because these properties hold for the cycle crossover $E_\chi$ and each valid $E_\chi$ corresponds to a unique offspring $c$.

A second, complementary offspring can be generated by repeating the last steps with the complementary choice $\chi^c=\{1,\dots,n\}\setminus\chi$. It is then necessary in step 5 to ensure that both offspring are full-length, because there is no guarantee that $E_\chi$ and $E_{\chi^c}$ have the same number of cycles.

Note that there are always at least two valid offspring, namely $c=a$ or $c=b$, corresponding to $\chi=\{\}$ and $\chi=\{1,\dots,n\}$, so the repetition in step $5$ never leads to an infinite loop.

\subsection*{Computational requirements for the asymmetric algorithm}

The proposed algorithm relies on a trial recombination which is repeated until a valid offspring is found. For assessing the computational requirements of the method, it is necessary to estimate how many trials are needed on average. We will investigate the number of trials required as a function of the problem size $n$ and the level of similarity between the parents. In addition, we will investigate how often the algorithm results in a trivial outcome, that is, an offspring that is identical to its parents, and therefore not a true recombination. While implementations can easily avoid these trivial cases in most cases, there are some pairs of parents for which no non-trivial recombination is possible, and we wish to avoid the bias of having to eliminate these cases. The frequency of trivial outcomes also serves as a rough measure of the variety of possible outcomes, since the outcomes that are identical to the parents will more rarely be sampled when there are many other possible recombinations.

\begin{figure}[tb!]
\begin{center}
\includegraphics[width=.5\linewidth]{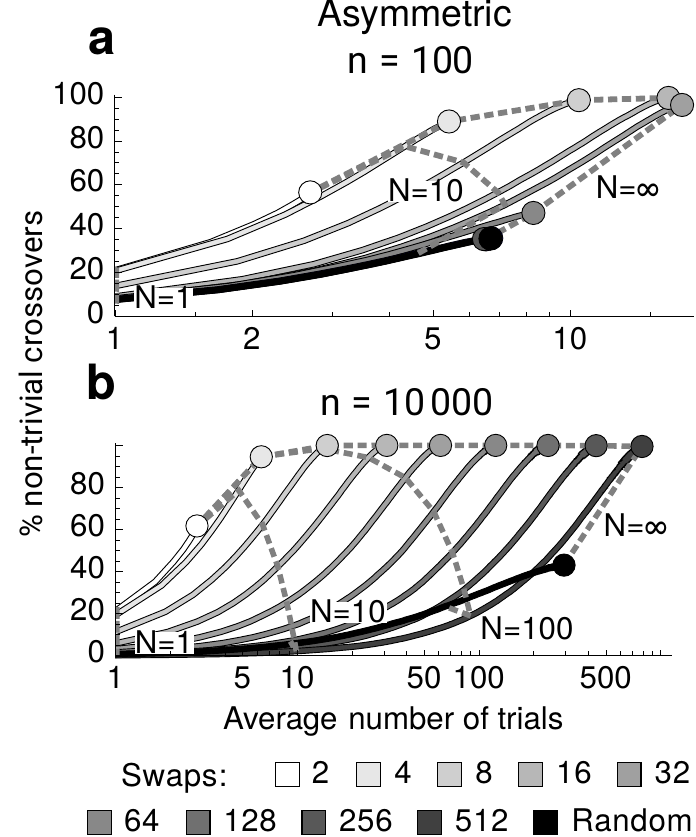}
\end{center}

\caption{Run time statistics of the proposed algorithm for (a) small permutations ($n=100$) and (b) large permutations ($n=10000$). Dots show the average number of trials (horizontal axis, logarithmic scale) and fraction of crossovers with non-trivial outcomes (vertical axis) for a constant level of divergence between the parents (shading, logarithmic scale). Solid curves show, for all possible values of $N$, how many crossovers have resulted in a non-trivial recombination after $N$ trials. Some specific values of $N$ are highlighted on the curves by dashed gray lines. Data shown are averages of 50000 crossovers.}
\label{fig:epxstats}
\end{figure}

Fig.~\ref{fig:epxstats} visualises these statistics for two different problem sizes. For each measurement, one parent was generated randomly, then mutated to produce a second parent by swapping a fixed number of random pairs. In the limit we have two random parents. Then, a crossover was performed, measuring the number of trials needed and whether or not the eventual outcome was distinct from its parents. To explore the distribution of the number of needed trials, we also show how many crossovers have finished (with a non-trivial outcome) after at most $N$ trials. While these measurements were made using random swaps of symbols, we expect similar results to hold for other kinds of parent dissimilarities with comparable disruption to the edges.

The number of trivial crossovers, in which the offspring is identical to one of the parents, is low in most scenarios, except when the parents are either highly similar or highly dissimilar. Since the offspring are uniformly sampled, a low probability of trivial outcomes indicates a high offspring variety. We observe that the offspring variety increases when there are more unique features in the parents to recombine, but decreases when the parents become so dissimilar that there is no common basis for recombination. Many pairs of random parents cannot be recombined at all, so that more than half of the sampled offspring are identical to their parents.

Not only the offspring variety increases with the parent dissimilarity, but also the expected number of trials. Interestingly, this increase follows a linear pattern. This pattern arises from the combination of two opposing trends, namely the simultaneous increase in the number of cycles (and thus the number of possibilities $\chi$ to choose from) and the number of valid offspring (and thus the number of choices $\chi$ that terminate the algorithm). The number of trials stops increasing when the parents are no longer similar enough to recombine effectively, and decreases asymptotically to the limit for random parents. The inability to recombine two parents that are not closely related can be compared to nature, where sexual reproduction is also impossible between individuals of very different species. 

At the transition between the linear and asymptotic regimes there is a ``worst-case'' level of similarity between the parents, where the number of trials is maximal. Unsurprisingly, this occurs when the number of swaps is a fixed fraction of the size ($y=0.266234x$, $r^2=0.99923$). Perhaps less intuitively, the expected number of trials at this maximum is likewise proportional to the problem size ($y=0.212657x$, $r^2=0.999816$). These linear regressions were performed on sample means taken from the distribution of trials, with sample size $50000$ for each data point. Problem size was varied between $100 \le n \le 1000$ in steps of $100$, then for each $n$ the highest mean was found in a set of populations with different levels of parent dissimilarity ($10 \le \text{}$swaps$\text{} \le 700$ in steps of $10$).

\section*{Asymmetric EAX performs CX on the adjacency representation}

Although derived differently, the algorithm described here is closely related to the asymmetric edge assembly crossover (EAX) algorithm. In fact, the first four steps of the proposed algorithm are mathematically identical to the first steps of the asymmetric EAX. In step 1, $E_a$ and $E_b$ are the adjacency-representations of the graphs $G_a$ and $G_b$. In step 2, the traversal of the union graph is done in steps following an edge from $G_b$ and then $G_a$ in reverse, or, in other words, following the edges of a graph with adjacency-representation $E_b E_a^{-1}$. The loops of even length in the graph are precisely the cycles of $E_b E_a^{-1}$ (where pairs of edges are represented in a cycle by their shared node). Note that the algorithm to generate AB-cycles in the asymmetric EAX is deterministic. Finally, in step 3 and 4, a union of cycles is exchanged to construct a recombined graph $G_c$ or adjacency-representation $E_\chi$. 

The two algorithms differ in the final step, or how they deal with invalid recombinations that contain more than one cycle ($E_\chi$) or connected component ($G_c$). In order to maintain perfect transmission, such invalid solutions must be discarded, and the algorithm proposed in the previous section simply repeats the process of recombination with a different set of cycles (i.e. E-set). In EAX, invalid solutions are considered ``intermediate'' and are patched by replacing low-quality edges (with respect to a given fitness function) with high-quality edges in a way that merges the disconnected cycles. During this process, the \textit{transmissive} property is lost. It also restricts the use of the EAX algorithm to the TSP, and not other applications where there is no known fitness function or its value is a non-linear function of the edges.

\section*{Transmissive crossover for undirected edges}

The asymmetric crossover described above can be described as a variant of asymmetric EAX in which the final step is replaced by repeated generation of intermediate solutions until a valid tour is found. The same approach can be applied to symmetric EAX, and must result in a \textit{transmissive} crossover for undirected edges.

Using the EAX algorithm to find AB-cycles, it is no longer possible to represent the exchange as a position-preserving crossover in the adjacency-representation. The recombination requires a more explicit exchange of edges. Note that also the orientation of edges may change, even for edges that are not part of the chosen E-set. When repeating the trial crossover, it is not generally necessary to repeat the procedure to identify different AB-cycles. We do so in our implementation in order to avoid the small source of bias that comes from setting an arbitrary rule for when to repeat the AB-cycle discovery procedure.

As described, this symmetric crossover variant is not \textit{respectful}: if the same edge is present in both parents (possibly in opposite orientation), it is not guaranteed that offspring also inherit this edge. Loss of the shared edge can occur when the two copies in the undirected union graph end up in a different AB-cycle, and one is in the E-set while the other is not. Luckily, this scenario can easily be avoided by removing both duplicates from the union graph before traversal. This does not affect the ability of the greedy algorithm to find a path, and ensures that the duplicate edges are not assigned to any AB-cycle, so that neither copy ends up in the E-set. 

We have not formally proven that this algorithm is able to produce all possible offspring adhering to the constraint of perfect transmission. Regardless, we note that it is not a \textit{perfect} crossover, because it does not sample its possible offspring in a uniform fashion, and so does not maximise the information entropy of the distribution of offspring. E-sets are drawn uniformly from all possibilities that result from a given traversal of the union graph, and the traversal is drawn uniformly as well, but different traversals can share the same AB-cycles and thus overlap in the E-sets that they can deliver. The algorithm is therefore more likely to produce offspring corresponding to E-sets that can be delivered by more different traversals. For example, an E-set composed of only one AB-cycle (or composed of the whole graph except a single AB-cycle) can be delivered by all traversals containing that cycle, whereas an E-set that is a union of many AB-cycles can only be delivered by traversals that contains all the relevant cycles. In the extreme, the trivial crossovers, which correspond to empty set and entire graph as E-sets and result in an offspring that is identical to one of its parents, can be delivered by any traversal of the graph, and are thus strongly over-represented in the outcome. Implementations may easily avoid the trivial E-sets.

\subsection*{Computational requirements for the symmetric algorithm}

\begin{figure}[tb!]
\begin{center}
\includegraphics[width=.5\linewidth]{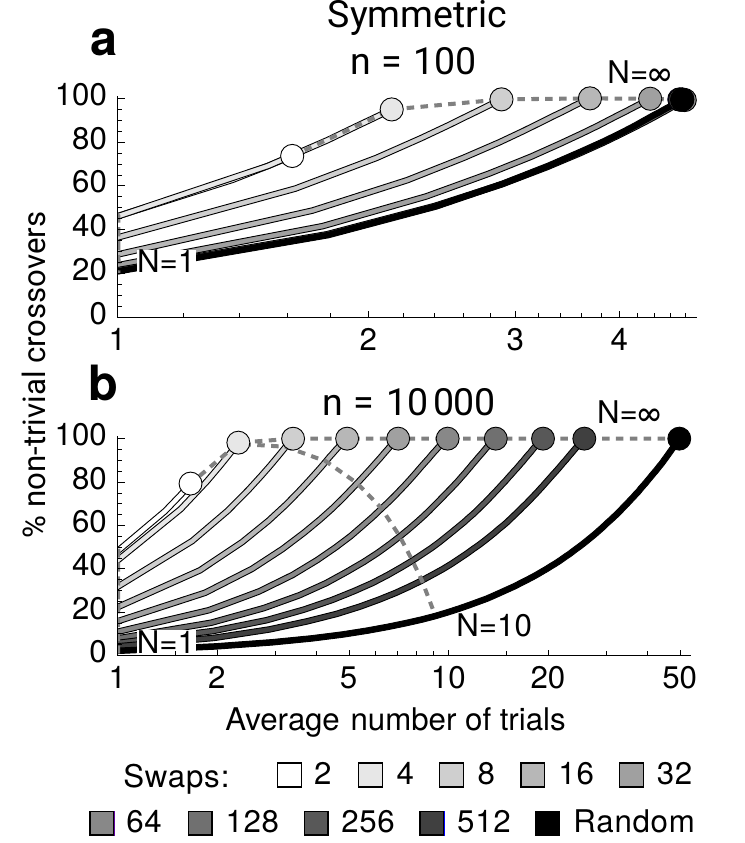}
\end{center}

\caption{Run time statistics of the symmetric variant of the proposed algorithm applied to (a) small permutations ($n=100$) and (b) large permutations ($n=10000$). }
\label{fig:sepxstats}
\end{figure}

Fig.~\ref{fig:sepxstats} shows the same runtime statistics as previously described, applied to the symmetric algorithm. Offspring which were equal to one of the parents but in opposite orientation were considered trivial crossover outcomes for these measurements. These data were gathered using the \textit{respectful} variant of the AB-cycle selection, where shared edges are taken out of the union graph before traversal.

There are remarkably few trivial crossover outcomes in these data. This is partially expected because the possible offspring of \textit{transmissive} crossover for undirected edges are a superset of those for directed edges. However, we concluded above that the trivial crossovers are strongly over-represented with this algorithm. Despite this, crossover outcomes were so diverse that no offspring was identical to one of its parents out of several populations of $50000$ samples, and in particular for all cases for the large problem $n=10000$ with more than $8$ swaps. We can use the number of AB-cycles to give a rough estimate of the number of offspring that these parents can generate. For example, at $16$ swaps, all recorded traversals of the union graph had at least $42$ AB-cycles, and the maximum number of trials needed was $51$; conservatively, this is evidence of at least $2^{42}/51\approx 10^{11}$ possible offspring. Notably, the asymmetric crossover can produce many possible offspring even from random parents. The number of trivial outcomes are shown in more detail in Table~\ref{tab:stats}.

Comparing the number of trials to the asymmetric crossover, we find that fewer trials are necessary. There is a similar proportional increase in the number of trials with the number of swaps, until the parents reach a certain level of dissimilarity. The expected number of trials is maximal when the swaps are a constant fraction of the problem size ($y=0.588571x$, $r^2=0.99731$) and the value at that maximum is also proportional to the size ($y=0.0263504x$, $r^2=0.994306$). In relation to the asymmetric crossover, the maximum occurs at twice the number of swaps and the maximal expected number of trials is eight times lower, approximately. 

\begin{table*}[htb!]
\caption{Number of crossover outcomes that were identical to one of the parents for problem size $n=10000$. These data are also shown in Fig.~\ref{fig:epxstats} and Fig.~\ref{fig:sepxstats} (vertical axis). Each cell is the number of trivial crossovers out of a batch size of $50 000$.}
\centering
\begin{tabular}{l|rrrrrrrrrr}
Swaps      & 2     & 4    & 8  & 16 & 32 & 64 & 128 & 256 & 512 & Random \\\hline
Asymmetric & 19148 & 2708 & 25 & 0  & 0  & 1  & 14  & 57  & 252 & 28461  \\
Symmetric  & 10358 & 924  & 11 & 0  & 0  & 0  & 0   & 0   & 0   & 0     
\end{tabular}
\label{tab:stats}
\end{table*}

\section*{Optimising crossover for the asymmetric TSP}

In case the fitness function is similar to the TSP, that is, the fitness is a linear combination of the choice of edges, it is possible to adapt the proposed algorithm to produce the offspring with the highest possible fitness, under the restriction that it contains only edges from its two parents. This approach only works for directed edges; for the undirected case, the offspring can only be optimised with respect to a particular set of AB-cycles. It is useful for applications where the quality of the offspring is more important than variety or computational efficiency. 

Assume a fitness function of the form $f_c=\sum_i^n \delta(c(i),c(i+1))$ for some offspring $c$, where $\delta$ defines the fitness of including an edge in a permutation. By rewriting this as $f_c=\sum_i^n \delta(i,E_c(i))$, we can associate a putative fitness value $f_\chi=\sum_i^n \delta(i,E_\chi(i))$ with any choice $\chi$, even if $E_\chi$ is not a valid edge transformed offspring. Expanding the definition of $E_\chi$, this can be further decomposed into $f_\chi=f_a(\chi)+f_b(\chi^c)$, where $f_a(S)=\sum_{i\in S} \delta(i,E_a(i))$ is the fitness cost of a set $S$ of edges in $a$, $f_b$ is the same for $b$, and $\chi^c=\{1,\dots,n\}\setminus\chi$ is the complement of $\chi$. Recalling that $\chi$ is defined as a union of cycles $k$, this can also be written as $f_\chi=\sum_{k\subseteq\chi} f_a(k) + \sum_{k\nsubseteq\chi} f_b(k)$. 

The union of cycles $\chi$ is defined by making a binary choice for each cycle $k$ of the successor mapping, and the corresponding fitness $f_\chi$ can be written as a sum of terms $f_a(k)$ or $f_b(k)$ depending on that choice. Using this decomposition it is possible to efficiently iterate over all possible choices $\chi$ (E-sets) in order of decreasing fitness, without listing all outcomes and ordering them, using an algorithm that iterates over subsets ordered by their sum. The first encountered E-set in this order that describes a valid offspring is then the edge-preserving crossover with the highest possible fitness. It is possible that the result is identical to one of the parents.


\subsection*{Computational requirements for the optimising algorithm}

\begin{figure}[tb!]
\begin{center}
\includegraphics[width=.5\linewidth]{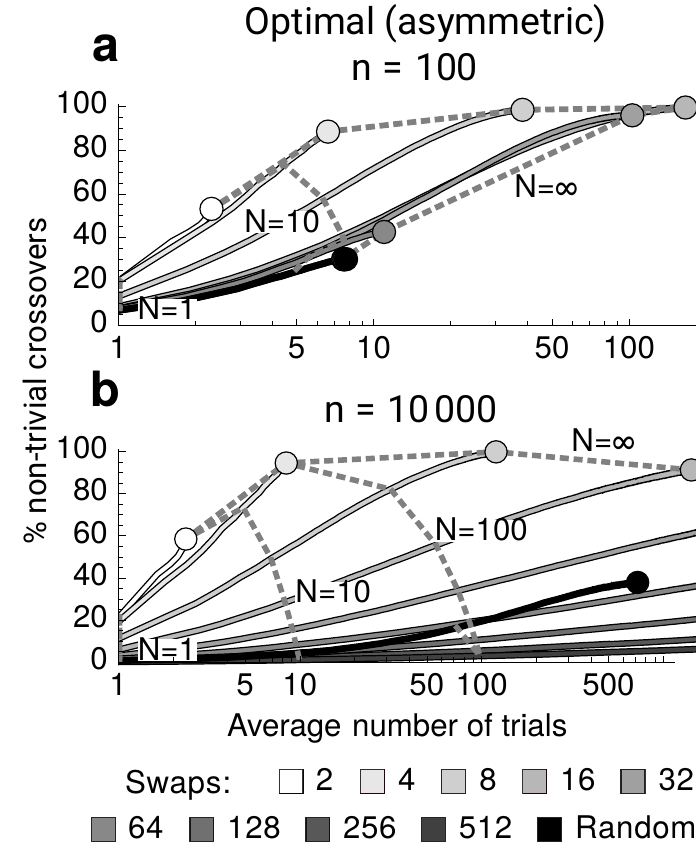}
\end{center}

\caption{Run time statistics of the EPOX algorithm for (a) a small problem ($n=100$) and (b) a large problem ($n=10000$).}
\label{fig:epoxstats}
\end{figure}

Runtime statistics, following the same motif as in the previous sections, are shown in Fig.~\ref{fig:epoxstats}. For these experiments we arbitrarily chose a TSP with Euclidian distances between cities on a rectangular grid. Since the initialisation of the parents is random, most pairs of parents have many possible offspring with higher fitness, resulting in a high percentage of non-trivial offspring. Realistic values in an evolutionary algorithm are likely to be much lower when selection has driven the population to a more optimised part of the fitness landscape, but this depends strongly on the algorithmic details. We will use these data to estimate algorithmic performance. 

The optimising crossover requires a much higher number of trials compared to the random crossover, especially if the parents are dissimilar. There is no linear proportionality between the number of swaps and the number of trials. This a consequence of the fact that the set of possible E-sets increases combinatorially when there are more unique features in the parents that can be recombined. In the random crossovers, this increase is offset by a concomitant growth in the number of allowed offspring, but the optimised crossover does not benefit from this because it searches for one particular offspring in all cases.

The computation times in the optimising crossover are skewed towards a relatively low number of crossover instances which require a relatively high number of trials. In Fig.~\ref{fig:epoxstats}, this is visible as a concave shape in the curves, showing that success of a trial becomes progressively less likely. The computational cost of these cases is very high for large genomes; some outliers required more than ten million trials for $n=10000$. For some applications it will be beneficial to limit to the number of trials to a value $N$, unless the problem size $n$ is low; the different values of $N$ shown in the figure correspond to the performance of the algorithm when the number of trials is limited to $N$ and the fall-back is counted as a trivial outcome. 

\section*{Conclusion}


Perfect transmission of parent edges to offspring is pervasive in literature as a gold standard for recombination since the concept of edge-based crossover was first published\cite{whitley_scheduling_1989}. On the other hand, crossover is only effective if it is capable of separating different features, or in other words produce a large variety of offspring. In some data structures, such as the bit strings that are traditional in genetic algorithms, it is possible to achieve both these goals exactly, in that it is possible to separate any combination of features and also retain error-free transmission. For the recombination of (directed or undirected) edges in permutations, however, this is not possible, so that correctness and variety are seen as a trade-off for crossover strategies\cite{radcliffe_algebra_1994,radcliffe_fitness_1995}.

However, existing crossover strategies emphasising correctness have not yet fully exploited that side of the trade-off, as they have not been able to produce the full variety of offspring under the constraint of \textit{transmissive} crossover. It is not encouraging that the problem is thought to be NP-hard\cite{moraglio_geometric_2011}, and some authors have speculated that errors in transmission are necessary to produce enough offspring variety and avoid premature convergence of the population\cite{nagata_criteria_2004}. We have shown here that \textit{transmissive} crossover is computationally tractable in practice, and able to produce a large variety of offspring.


In line with the suspected NP-hardness of the problem, the algorithms presented here have a worst-case asymptotic complexity that is slower than polynomial time. For example, in the asymmetric algorithm, the complexity depends on the number of independent cycles that can be inherited by the offspring. For a problem size of $n$, we may have up to $n/2$ non-singleton cycles, and therefore we may have to attempt up to $2^n-1$ trials before finding one of the two only guaranteed possible outcomes. According to this naive analysis, the algorithm is $O(n 2^n)$, since each trial has a linear computational cost. However, the expected computational complexity is more informative for practical run times than the worst case. While a permutation of size $n$ may have any number of cycles between $1$ and $n/2$, the vast majority of permutations have only a few cycles, with an average that scales as $\log(n)$. Based on this average, we may issue a naive expectation that the number of trials scales linearly, as opposed to exponentially, as a function of problem size. A similar estimation can be obtained by considering that a fraction $1/n$ of crossovers with size $n$ have only one cycle, so that an average of $n$ trials would be necessary to find a valid adjacency-representation by cycle crossover if the outcome is approximated as random.

Here we have shown, perhaps surprisingly, that this is indeed the case. The expected number of trials is best understood as a function of the dissimilarity $s$ between the parents, and not only of the problem size. However, for a given problem size $n$, the expected number of trials as a function of $s$ ($\text{trials}_n(s)$) has a maximum value proportional to $n$. This proportionality implies an expected $O(n^2)$ computation time for both the symmetric and asymmetric random crossover strategies, in the ``worst case'' where the crossover usually occurs between parents with a level of dissimilarity near the maximum. In practice, it is rare to recombine such dissimilar parents in evolutionary algorithms. For low (and more realistic) values of $s$, the expected $\text{trials}_n(s)$ is a linear function, with a slope independent of $n$. Genetic algorithms dealing with larger problems tend to have more population variation in terms of absolute number of differences, so the practical cost of crossover in this setting is expected to be between $O(n)$ and $O(n^2)$ depending on the specific application. 


The proposed algorithm is not only unexpectedly efficient, but also produces a striking amount of variety. We have shown that the number of possible offspring that can be generated by symmetric \textit{transmissive} crossover explodes combinatorially with the number of unique features in the parents. In our results, we have restricted the offspring to retain all edges that occur in both parents; without this restriction, which is an optional part of the algorithm, an even higher variety can be generated. For asymmetric recombination, trivial crossover outcomes were also rare for parents with dissimilarity levels that are realistic for situations where genetic recombination is likely to be applied. Based on these observations, we believe that the trade-off between correctness and variety is not as stringent as previously indicated\cite{nagata_edge_1997,nagata_criteria_2004}. 

The derivation of our algorithm reveals a close connection with two other, seemingly unrelated crossover strategies. The first four steps are identical to cycle crossover performed on the adjacency-representation, and also to the construction of an intermediate solution in the asymmetric EAX. The insight that these two are the same is new, and arguably simplifies the asymmetric EAX. This formulation is closer to implementation, since cycle cross and transformation between path- and adjacency-representation are both simple operations on integer arrays. Conversely, the EAX is a mature algorithm supported by an academic history that has produced some interesting insights and variations, some of which can also be applied to improve the proposed algorithm for \textit{transmissive} crossover. For example, Nagata et al.\ have investigated several different ways to construct an E-set given a set of AB-cycle candidates. EAX-1AB\cite{nagata_edge_1997} and EAX-block\cite{nagata_new_2006} modify the distribution of E-cycles by choosing only one AB-cycle or biasing the selection towards groups of AB-cycles that are close, respectively, in order to minimise the expected number of cycles in the intermediate solution. The same techniques should result in a lower number of trials for \textit{transmissive} crossover, in either the symmetric or asymmetric case, although at the cost of biasing the distribution of offspring and thus decreasing the effective variety.

While previous modifications of EAX have focused on reducing computation time, some applications would likely benefit from an improved variety of offspring. In future work, it would be useful to formally prove that the proposed algorithm can generate any possible offspring consisting only of edges from the parents; or, if this proves not to be the case, provide an alternative algorithm that can. In addition, the effective variety of the offspring is diminished due to a bias in the procedure to find AB-cycles. It may be possible to reduce this bias, or perhaps even eliminate it, by appropriately modifying the selection of E-sets.

\printbibliography

\section*{Acknowledgements}

The authors thank Fiona Mulvey for extensive proofreading of the manuscript.

\end{document}